\def\year{2019}\relax
\documentclass[letterpaper]{article} 
\usepackage{aaai19}  
\usepackage{times}  
\usepackage{amsfonts}
\usepackage{amsmath}
\usepackage{helvet}  
\usepackage{courier}  
\usepackage{url}  
\usepackage{graphicx, subfigure}  
\usepackage{xcolor} 
\usepackage{makecell} 
\usepackage{pgfgantt}
\usepackage{anyfontsize}
\usepackage{multicol}
\begin{document}
%
\title{Constructing Hierarchical Q\&A Datasets for Video Story Understanding} 

\setlength\titlebox{4.2in}
\author{Yu-Jung Heo, Kyoung-Woon On, Seongho Choi\\Department of Computer Science and Engineering\\Seoul National University\\Seoul, South Korea\\ \{yjheo, kwon, shchoi\}@bi.snu.ac.kr
\And Jaeseo Lim\\Interdiscipliary Program in Cognitive Science\\Seoul National University\\Seoul, South Korea \\ jaeseolim@snu.ac.kr
\AND Jinah Kim \\ Marine Disaster Research Center \\ Korea Institute of Ocean Science \& Technology(KIOST) \\ Busan, South Korea \\ jakim@kiost.ac.kr
\And Jeh-Kwang Ryu \\ Institute for Cognitive Science \\ Interdisciplinary Program in Cognitive Science \\ Seoul National University \\ Seoul, South Korea \\ ryujk@snu.ac.kr
\AND Byung-Chull Bae\\School of Games\\Hongik University\\Sejong, South Korea \\ byuc@hongik.ac.kr
\And Byoung-Tak Zhang\\Department of Computer Science and Engineering\\Seoul National University\\Seoul, South Korea \\ btzhang@bi.snu.ac.kr
}

\maketitle
\begin{abstract}
Video understanding is emerging as a new paradigm for studying human-like AI. Question-and-Answering (Q\&A) is used as a general benchmark to measure the level of intelligence for video understanding. While several previous studies have suggested datasets for video Q\&A tasks, they did not really incorporate story-level understanding, resulting in highly-biased and lack of variance in degree of question difficulty. In this paper, we propose a hierarchical method for building Q\&A datasets, i.e. hierarchical difficulty levels. We introduce three criteria for video story understanding, i.e. memory capacity, logical complexity, and DIKW (Data-Information-Knowledge-Wisdom) pyramid. We discuss how three-dimensional map constructed from these criteria can be used as a metric for evaluating the levels of intelligence relating to video story understanding.

\end{abstract}

\section{Introduction}
In narratology, story is often differentiated from discourse, where story refers to content (i.e., what to tell) and discourse denotes expression or representation (i.e., how to tell it)\cite{Chatman1978,Genette1980}. While the representation can vary from text and oral storytelling to films, dramas, and virtual environments including virtual reality (VR), understanding of the given story shares some common key aspects regardless of the represented media.

According to computational linguists, narrative theorists, and cognitive scientists, narrative understanding is somehow linked with the measurement of reader's intelligence. For example, readers can understand story as a way of problem solving in which they keep focusing on how main characters overcome coming obstacles throughout story \cite{Black1980}. Thus readers, while reading, make inferences both in prospect and in retrospect about what events will occur and how these events could occur, considering the causal relationships between different events in the story \cite{trabassovandenbroek1985causal,McKoon1992,Graesser:1994}. Inferring causal relationships between events, is a key element for the reader to reconstruct a given narrative as a mental model in the reader's mind \cite{Zwaan1999,Zwaan1995}. Humans have the natural capability of ``organizing our experience into narrative form" as narrative intelligence \cite{Blair1997,mateas1999}. 

Recently, video story serves as a testbed of real-world data to construct human-level AI from two points of view. First, video data has various modalities such as sequence of images, audios (including dialogue, sound effects, background music), and often texts (subtitles or added comments). Second, the video shows a cross-section of everyday life. Understanding video story involves analyzing and simulating human vision, language, thinking, and behavior, which is a significant challenge to current machine learning technology.

To measure human-level machine intelligence, we apply video Question-and-Answering (video Q\&A) task as a proxy of video story understanding. The task can be regarded as a Turing Test for video story understanding\cite{turing1950computing}. While several previous studies have suggested various datasets for the video Q\&A task\cite{Tapaswi2015,Kim2017,mun2017marioQA,Jang2017,lei2018tvqa}, they are built without careful consideration of ``understanding of video story''. For such reason, the previously released video Q\&A datasets are highly-biased and lack of variance in question difficulty. The construction of Q\&A dataset with hierarchical difficulty levels in terms of story understanding is crucial, as people with different perspectives (or different intelligence levels) will understand the given video story differently.   

In this paper we propose three criteria such as memory capacity, logical complexity, and DIKW hierarchy for video story understanding and construct a three-dimensional hierarchical map of video story understanding using the criteria. The constructed three-dimensional map can leverage the understanding of developmental stages of human intelligence. We expect that the proposed hierarchical criteria can be utilized later as a metric that can help evaluate the levels of intelligence relating to video story understanding. Our main contributions are twofold. First, we suggest three criteria for constructing hierarchical video Q\&A datasets. The criteria can be used to analyze the quality of video Q\&A dataset in terms of bias and variance for dataset difficulty. Second, we interlink proposed three criteria to neo-Piagetian's theory, which can help interpreting our story-enabled intelligence to cognitive development stage of human.

\section {Related Works} 

While video understanding is still in its early stage, researchers proposed video Question-and-Answering (video Q\&A) dataset as a general benchmark to measure video understanding intelligence. The most notable datasets proposed so far are MovieQA\cite{Tapaswi2015}, PororoQA\cite{Kim2017}, MarioQA\cite{mun2017marioQA}, TGIF-QA\cite{Jang2017}, and TVQA\cite{lei2018tvqa}. Here, we review above video Q\&A datasets and present our contributions.

MovieQA aims to evaluate story understanding of video and text in movie. For the MovieQA, question and answer pairs are collected by annotators who read plot synopses of movies instead of the entire movies. PororoQA is comprised of targeted animation videos, which makes its content easier to understand than MovieQA dataset. MarioQA dataset is also  based on synthetic videos constructed automatically from the popular Mario game playing videos. The dataset focuses on understanding of temporal relationship between multiple events. When the dataset is generated, template-based question and answer generation methods are used from extracted events. TGIF-QA dataset focuses on only visual information in the GIF-format images. TGIF-QA limits the question type to three types: repetition count, repeating action, and state transition. Those types of questions are required spatio-temporal reasoning from videos. TVQA question is a large-scale video QA dataset based on 6 popular TV shows about sitcoms, medical and crime TV programs. For the TVQA, all questions and answers are attached to 60-90 seconds video clips. It requires comprehension for subtitle-based dialogue and recognition of relevant visual concepts to answer the questions in the dataset.

Our work contributes to this line of research, but instead of introducing new datasets, here we propose new criteria for constructing video Q\&A dataset on careful consideration of video story understanding.

\section{Three-dimensional Video Q\&A Hierarchy} \label{problem-definition}
This section describes three criteria as measures of video story understanding. The three criteria are as follows: memory capacity, logical complexity, and DIKW (Data-Information-Knowledge-Wisdom) hierarchy. These three criteria are combined to construct a three-dimensional video understanding map. Every question in the Q\&A dataset is classified in each level by each criterion respectively, and then represented as a point on the three-dimensional map which has three dimensions corresponding to the three criteria. Finally, every point in the map is assigned to the cognitive development stage according to Piaget's theory\cite{piaget1972intellectual,collis1975study}. We explain this process in detail in the following subsections.

\subsection{Criterion 1: Memory Capacity}
When determining the difficulty of questions collected for the video, the length of the video is crucial for reasoning and finding the correct answer in machine learning perspective. If the length of the video required for answering the question is longer, the question can be classified as more difficult, and vice versa. For example, a question targeted to short video is a lot more difficult than a question targeted to an image frame, and a question targeted to entire video is a lot more difficult than a question targeted to one segment video. This criterion also can be interpreted as memory capacity of humans. In this paper, We define \textit{Memory Capacity} as the length of the target video which has to be considered to answer given question. We use the terms defined at each level consistently with the terms in \cite{zhai2006}. The classification results are as follows.
 
\begin{itemize}
    \item Level 1 (frame): The questions for this level are based on a video frame. This level has the same difficulty as that of a kind of Visual Question Answering (VQA) dataset\cite{malinowski2015ask,ren2015exploring,Agrawal2017,zhu2016visual7w,johnson2017clevr,wang2018fvqa}.
    
    \item Level 2 (shot): The questions for this level are based on a video length less than about 10 seconds without change of viewpoint. This set of questions can contain atomic or functional/meaningful action in the video. Most recent datasets which deal with video belong to this level\cite{Jang2017,Maharaj2017,mun2017marioQA}. The questions of this level aim to evaluate understanding of information which contains video characteristics that are not in Level 1 (frame level). One important point is that target video for questions at this level contains meaningful actions. At this level, both atomic action and meaningful action can appear, and their boundary is vague. For example, waving hands (atomic action) and a gesture to saying goodbye (meaningful action) have a similar action, However, their meaning is different depending on the situation, not depending on video length. This level contains both of actions, even if their difficulty is clearly different.
    
    \item Level 3 (scene): The set of questions for this level is based on clips that are 1-3 minutes long without place change. Videos at this level contain sequences of actions, which augment the level of difficulty from Level 2. We consider this level as the ``story'' level according to our working definition of story. MovieQA\cite{Tapaswi2015} and TVQA\cite{lei2018tvqa} are the only datasets which belong to this level. For example, the popular TV sitcom \textit{Friends} has 13 scenes per episode on average, and a movie has 120 scenes on average.

    \item Level 4 (sequence): The set of questions at this level is related to more than two scenes, but less than entire movie. To the best of our knowledge, there are no datasets dealing with the video at this level.
    
    \item Level 5 (entire): The question set for this level is based on an entire story from beginning to end. Questions at this level are based on whole video such as an entire movie or an episode of a drama.

\end{itemize}

\subsection{Criteria 2: Logical Complexity}
Complicated questions often require more (or higher) logical reasoning steps than simple questions. In other words, if a question requires multiple supporting facts which have interrelations to answer, we regard that the question has high logical complexity. For story-enabled intelligence, it is required to trace several logical reasoning steps by combining multiple supporting facts to give a correct answer to a given question. In a similar vein, if a question needs only a single supporting fact with a single relevant datum, we regard that it has low logical complexity. It may need only one reasoning step or one perception step to answer the question. 

This subsection describes the second criterion \textit{logical complexity} to define the level of difficulties for questions. We define five logical complexity levels based on the Stanford Mobile Inquiry-based Learning Environment(SMILE)\cite{seol2011stanford}. In the SMILE project, students learn online lectures or documents via a mobile platform and generate relevant questions based on what they have learned. Each question made by a student is classified into five logical complexity levels as follows:

\begin{itemize}
    \item Level 1 (Simple recall on one cue) : The question set at this level can be responded with minimal cognitive effort, involving simple recall or simple arithmetic calculations. The questions at this level require only one supporting fact. Supporting fact is a triplet form of \textit{\{subject-relationship-object\}} such as \textit{\{person-hold-cup\}}. As the questions at this level are too simple, they may not trigger much interaction.
    
    \item Level 2 (Simple analysis on multiple cues) : The question set at this level can be responded with simple analysis of the question types or problems with simple reasoning. The questions at this level ask for factual information involving recall of independent multiple supporting facts, which trigger simple inference or quick interpretation. For example, two supporting facts \textit{\{tom-in-kitchen\}} and \textit{\{tom-grab-tissue\}} are referenced to answer ``Where does Tom grab the tissue?". This question set begins with simple question types starting with ``Who", ``What", ``When", ``Where", ``How many", and so on. Responses come from a range of clearly defined scope with little room for dispute.

    \item Level 3 (Intermediate cognition on dependent multiple cues) : The question set can be responded with intermediate level of cognition and analysis. The questions at this level require multiple supporting facts with time factor. Time factor is a sequence of the situations or actions. Accordingly, the questions at this level cover how situations have changed and subjects have acted. It also requires cognitive operations such as comparison, classification, or categorization in responding to given questions at this level.

    \item Level 4 (High-level reasoning for causality) : The question set at this level can be responded with higher-level of analysis and reasoning rather than a lower-level thinking question. The question set covers reasoning for causality beginning with ``Why''. Reasoning for causality is the process of identifying causality, which is the relationship between cause and effect from actions or situations. It requires own interpretation or synthesis in responding to given questions at this level.
    
    \item Level 5 (Creative thinking) : The question set at this level can be responded by requiring imagination and creation of new theory or hypothesis with supporting rationale. The question at this level covers creative thinking and reasoning that may help defining a new solution or concept that has not existed previously. For example, the questions (\#19 and \#20) in Table 1 on appendix draw an unique solution by formulating own rational equations about not occurred situations.
    
\end{itemize}

\subsection{Criterion 3: DIKW Hierarchy}
The DIKW (Data, Information, Knowledge, and Wisdom) hierarchy is widely accepted as a way of representing different levels of what we see and what we know \cite{schumaker2011data}. 
In terms of video Q\&A, a level of understanding can be also identified by answering questions based on different levels ranging from data, information, knowledge, and wisdom. In this section, details of the four levels of video understanding are discussed.
 
\begin{itemize}
    \item Level 1 (Data-level): Data are the observations of the physical world \cite{schumaker2011data,carlisle2006escaping} and are symbolic representation of things, events and activities\cite{ackoff1989data,rowley2007wisdom}. In terms of video Q\&A, the data-level covers the questions for characters, characters' lines, objects, sounds, locations and simple behaviors which have no subjective meaning or goal for the environment (such as standing, walking, calling, and etc.).
    
    \item Level 2 (Information-level) : Information refers to the data that have been shaped into a meaningful and useful form \cite{rowley2007wisdom}. Specifically, it includes the addition of relationships between data \cite{barlas2005self}. In terms of video Q\&A, information-level questions focus on the meaningful interaction between characters and objects such as actions, emotions, and situations that can be obtained from the scene of the video.
    
    \item Level 3 (Knowledge-level) : Knowledge refers to the aggregation of related information that provides a clear understanding of information \cite{barlas2005self,schumaker2011data}. Knowledge also involves the synthesis of multiple sources of information over time \cite{rowley2007wisdom,despres2012knowledge}. In terms of video Q\&A, knowledge-level questions can be answered only with accumulated information of  multiple scenes of the video including knowledge from fictional universe of contents and commonsense.
    
   \item Level 4 (Wisdom-level): Wisdom refers to accumulated knowledge, with which it is possible to apply understood concepts from one domain to new situations or problems \cite{rowley2007wisdom}. In terms of video Q\&A, the wisdom-level questions can be answered by utilizing useful meta relationship of knowledge including nonsense and humor. For example, the question \#28 in Table 1 on appendix needs to understand the character ``Chandler" in terms of a sense of humor.
   
\end{itemize}

\begin{figure*}[t]
\begin{center}
\includegraphics[width=0.85\textwidth]{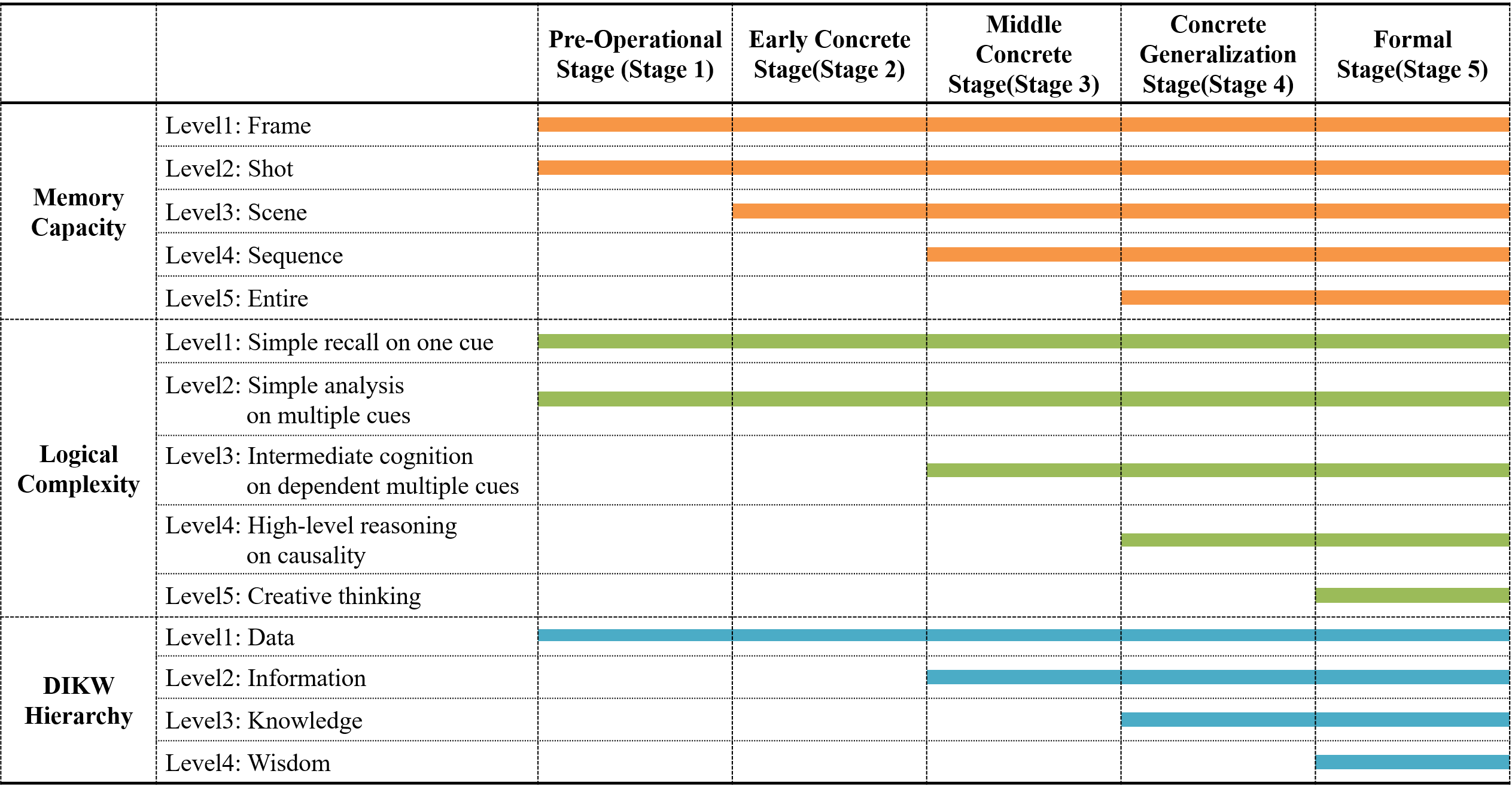}
\end{center}
   \caption{Interpretation of the proposed three criteria \textit{(i.e., Memory capacity, Logical complexity, and DIKW Hierarchy)} as cognitive development stage proposed by Piaget and recasted by Collis. The highlighted bar means the possibility to apply cognitive operations for answering given question on each level of a criterion from each cognitive developmental stage.}
\label{fig:aaaisss_fig1}
\end{figure*}

\subsection{Interpretation as Cognitive Development Stage}
In the following section, we interpret proposed three criteria \textit{(i.e., memory capacity, logical complexity, and DIKW pyramid)} from the viewpoint of cognitive development of human intelligence. The detailed cognitive development defined by Piaget is introduced, and then we apply the cognitive development stage to criteria of three-dimensional video Q\&A hierarchy.

\subsubsection{Piaget's theory of cognitive development}
In this section, we explain cognitive development of human based on one of Neo-Piagetian theory\cite{collis1975study} recasting of Piaget's theory of developmental stages\cite{piaget1972intellectual}. Piaget's theory explains in detail the process by which human cognitive ability develops, in conjunction with information processing models. In order to justify three criteria proposed in this paper in terms of human intelligence development, we examine the details of the developmental stages of Piaget's theory. Piaget's original model suggests a sensory-motor stage that occurs from birth, however, that stage involves only representations related to sensory-motor activity. Thus, we focus on the later stages that follow the pre-operational stage in which a child shows understanding behavior\cite{collis1975development}.

\begin{itemize}
    \item Stage 1 (Pre-Operational Stage; 4 to 6 years) : At this stage, a child thinks at a symbolic level, but is not yet using cognitive operations. The child can not transform, combine or separate ideas. Thinking at this stage is not logical and often unreasonable. Associations are made on the basis of emotion and preference at this stage, and it has a very egocentric sight of one's own world. 
	
	\item Stage 2 (Early Concrete Stage; 7 to 9 years) : At this stage, a child can utilize only one relevant operation. Thinking at this stage has become detached from instant impressions and is structured around a single mental operation, which is a first step towards logical thinking.
	
	\item Stage 3 (Middle Concrete Stage; 10 to 12 years) : At this stage, a child can think by utilizing more than two relevant cognitive operations and acquire the facts of dialogues. This is regarded as the foundation of proper logical functioning. However, a child at this stage lacks own ability to identify general fact that integrates relevant facts into coherent one. Moreover, thinking at this stage is still concrete, not abstract.
    
    \item Stage 4 (Concrete Generalization Stage; 13 to 15 years) : Piaget referred to this stage as the early formal stage, particularly for abstract thinking. A child at this stage, however, can just generalize only from personal and concrete experiences. The child do not have own ability to hypothesize possible concepts or knowledge that is quite abstract.
	
	 \item Stage 5 (Formal Stage; 16 years onward) : This stage is characterized purely by abstract thought. Rules can be integrated to obtain novel results that are beyond the individual’s own personal experiences. However, this is not a stage that every person can reach.
\end{itemize}

\begin{figure*}[t]
\begin{center}
\includegraphics[width=0.90\textwidth]{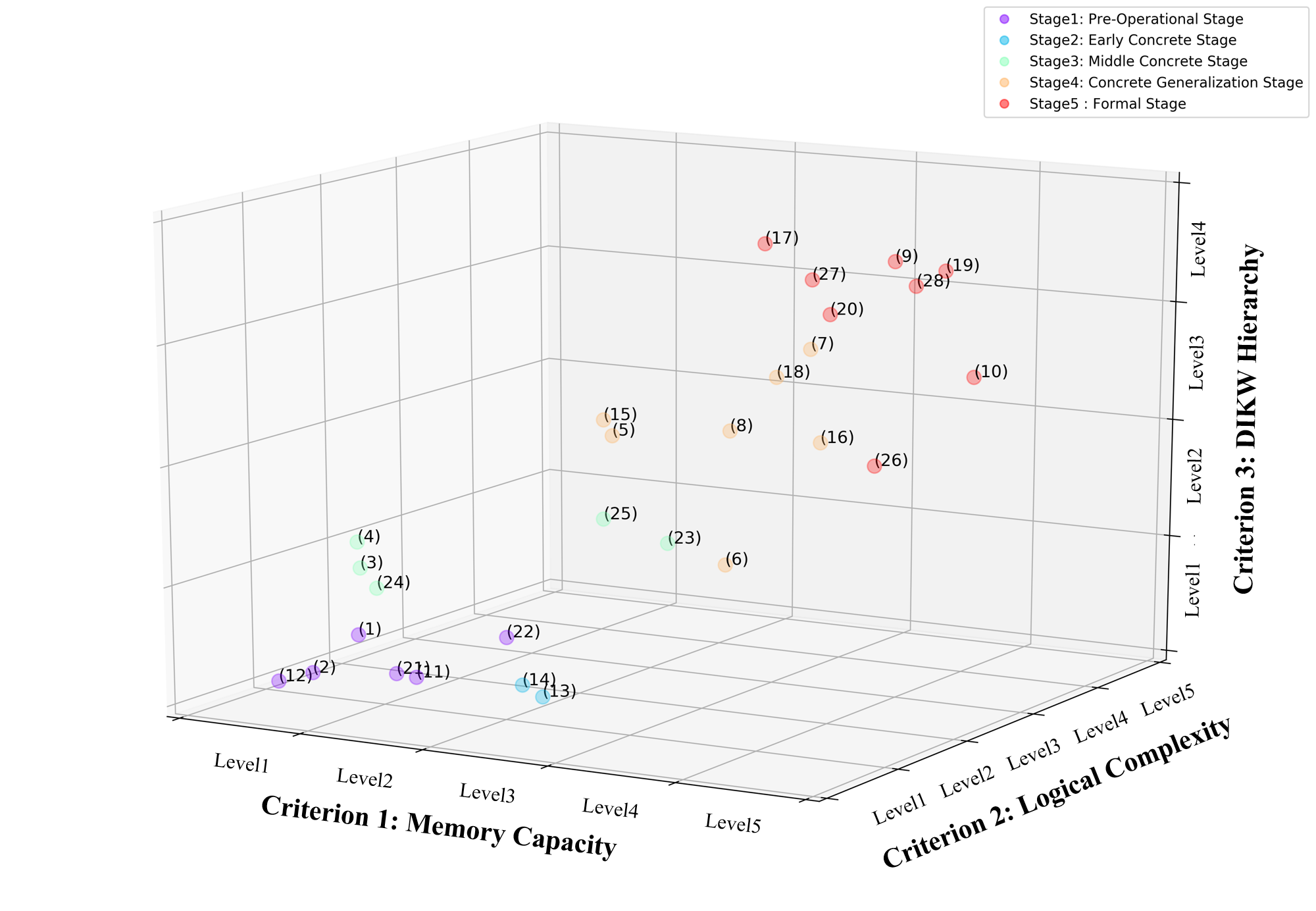}
\end{center}
   \caption{Example of three-dimensional video Question-and-Answering (video Q\&A) hierarchy. Each point represents each question in Table 1 on appendix. Three coordinates of each point is assigned by the definition of level of three criteria, and matched to one developmental stage which is the highest stage among derived three stages following the interpretation of Figure \ref{fig:aaaisss_fig1}.}
\label{fig:aaaisss_fig2}
\end{figure*}

\subsubsection{Applying Piaget's human developmental stage to criteria of three-dimensional hierarchy}

Human understanding, as Piaget stated, can be classified into different stages. We propose that the concepts from Piaget's theory of development correspond to the three-dimensional video Q\&A Hierarchy criteria.

First, the development stage can be explained from the perspective of the memory capacity criterion. Memory capacity corresponds to working memory of the cognitive process model. \cite{case1980implications} suggested that the working memory available for problems increases with age, as does the space required for higher level responses. This relationship between working memory and age leads to the proposition that cognitive developmental stages can be explained by increasing attention span, or working memory capacity\cite{case1980underlying,mclaughlin1963psycho,pascual1969cognitive}. Thus, we assume that Piaget's theory of development of human intelligence with age can be in accordance with the memory capacity criterion. For example, understanding a static image can be understood to be from Stage 1 (Pre-operational). Also, understanding video within 10 seconds is possible from Stage 1. However, beyond minutes\textit{(e.g., understanding a scene within 3 minutes)}, is possible from Stage 2 (Early concrete). Beyond this, understanding two or more scenes \textit{(e.g., understanding sequences changing time and place)} is possible from Stage 3 (Middle concrete). Finally, it is possible from Stage 4 (Concrete generalization) to know and understand whole video entirely. While this mapping is not exactly distinct, there exists a clear hierarchy as shown in Figure \ref{fig:aaaisss_fig1}.

Piaget's developmental stages are also consistent with the logical complexity criterion. As the SMILE project proposes, from simple recall to assumption-based reasoning, methods have a kind of hierarchy that is closely related to a person's stage of development. For example, level 1 and level 2 is available from Stage 1 (Pre-operational) in that it needs a simple call. Specifically, level 1 requires one supporting fact \textit{(e.g., \{jacket-is-black\})}, on the other hand, level 2 requires independent multiple supporting facts. Level 3 is available from Stage 3 (Middle concrete), in that this level can be understood using dependent multiple supporting facts across time. Level 4 is available from Stage 4 (Concrete generalization), because this level requires a higher thought on causality in relation to ``Why''. Finally, level 5 is available from Stage 5 (Formal Stage), as it requires creativity and abstract thinking about new ideas. As such, each phase of SMILE can be expressed as roughly equivalent to the human developmental stage postulated by Piaget\cite{collis1972concrete}.

Piaget's human developmental stages are not exactly consistent with the DIKW hierarchy criterion. However, data-level is possible from the Stage 1 (Pre-operational) in terms of providing simple factual data. Information-level is possible from Stage 3 (Middle concrete), because it identifies a relationship between some real world entities. Knowledge-level is roughly equivalent from Stage 4 (Concrete generalization), as both are incapable of inferring information from abstract variables. Finally, wisdom-level in DIKW hierarchy criterion is possible from Stage 5 (Formal stage), in that it can be inferred and applied to new situations like knowledge transferring\cite{collis1975development}. Figure \ref{fig:aaaisss_fig2} shows three-dimensional hierarchical map for each question represented in Table 1 on appendix. Three coordinates of each point are assigned by the definition of level of three criteria, and matched to one developmental stage which is the highest stage among derived three stages following the interpretation of Figure \ref{fig:aaaisss_fig1}. For example, a question ``What did Joey and Chandler do at Ross house?" is set \{3, 2, 1\} level for three criteria. Each level can be developed from \{2, 1, 1\} stage, so that the question is mapped Stage 2 which is the highest stage among \{2, 1, 1\} stage.

\section{Discussion and Future Works}
In this paper, we propose a theoretical framework with three criteria \textit{(i.e., memory capacity, logical complexity, and DIKW hierarchy)} to construct a hierarchical Q\&A dataset for video story understanding. A key contribution of our work is to suggest an approach to classify the difficulties of questions for video story understanding in accordance with three criteria. Interestingly, the three criteria can be mapped with the five stages of human development in Piaget's theory, which can serve as a basis for Q\&A systems to evaluate video story understanding.

While suggested three criteria can be linked to human developmental stages, it is hard to define separately and exactly. This is due to the limitation of Piaget's theory, where the human development stage cannot be discontinuous. That is, each developmental stage is not always precisely differentiated by age and cognitive abilities. Especially, the agreement for development stage of the high-level cognition is still controversial. Nevertheless, Collis' stages\cite{collis1972concrete,collis1975development,collis1975study} are the appropriate attempt to classify understanding according to human development stages. Furthermore, the connection from the three criteria to five developmental stages presents the possibility that our story-enabled intelligence can be associated to cognitive development stages of human. Applying knowledge about human cognitive development will help to set the direction of human-level AI research in detail.

Moreover, comparing with human development stages, machine learning approach requires clear and explicit specification. For example, in Collis' classification, each stage has an approximate two-year interval, while the machine needs to be viewed with more detailed and specific classification criterion - such as either every month or every season.

As future work, we plan to extend the proposed criteria to reflect some viewpoints from cognitive narratology (e.g., Zwaan's five index model of narrative understanding including space, time, characters, goals, and causation\cite{Zwaan1999}). We also plan to use our proposed criteria as a guideline for video story understanding to construct a carefully designed hierarchical dataset. The proposed video Q\&A hierarchy can be used as a metric for the developmental level of machine intelligence, and as a guidance to what dataset should be collected to study the desired level of machine intelligence.

\subsubsection{Acknowledgments.}
The authors would like to thank Woosuk Choi and Chris Hickey for helpful comments and editing. This work was partly supported by the Korea government (No.2017-0-01772, Development of QA systems for Video Story Understanding to pass the Video Turing Test and IITP-R0126-16-1072-SW.StarLab, 2018-0-00622-RMI, KEIT-10060086-RISF, NRF-2016R1D1A1B03936326)

\bibliographystyle{aaai}
\bibliography{aaai19}

\newpage
\setlength{\tabcolsep}{6pt}
\begin{table*}\label{table:examples} \section{Appendix. }
\begin{center}
\caption{Examples of Question-and-Answering datasets for each criterion based on TV sitcom Friends}
\begin{tabular}{ll}
\noalign{\smallskip}
\Xhline{3\arrayrulewidth}
\multicolumn{2}{l}{\textbf{Criterion 1: Memory capacity}}\\
\hline
Level 1: Frame
& (1) Q: What does Ross have on his shoulder?\\& \hspace{0.42cm} A: A monkey is on his shoulder.\\
& (2) Q: What is the hat made out of? \\& \hspace{0.42cm} A:The hat is made out of a cup.\\
\hline
Level 2: Shot
& (3) Q: What did Chandler swallow? \\& \hspace{0.42cm} A: The button off his shirt.\\
& (4) Q: How many times did the Ross knock the door? \\& \hspace{0.42cm} A: Five times. \\

\hline
Level 3: Scene
& (5) Q: What did Rachel do wrong when she did her first load of laundry? \\& \hspace{0.42cm} A: Rachel left a red sock in her load of white clothes which made them pink.\\
& (6) Q: Why does Rachel's boyfriend visit Phoebe at work? \\& \hspace{0.42cm} A: To get a massage.\\ 
\hline
Level 4: Sequence
& (7) Q: Why does Rachel not want to drink on her date? \\& \hspace{0.42cm} A: Because she doesn't want to end up in his bed on the first date.\\
& (8) Q: Why is Ross having the locks replaced? \\& \hspace{0.42cm} A: He lost all of his keys. \\
\hline
Level 5: Entire
& (9) Q: What is the main topic of this movie? \\& \hspace{0.42cm} A: The story about true love.\\
& (10) Q: How does the main character's personality change from the beginning to end? \\& \hspace{0.57cm} A: At first he was fainthearted, but later he became brave.\\

\hline

\Xhline{3\arrayrulewidth}
\multicolumn{2}{l}{\textbf{Criterion 2: Logical Complexity}}\\
\noalign{\smallskip}
\hline
Level 1: Simple recall 
& (11) Q: Who is drinking water?\\ & \hspace{0.57cm}   A: Monica is drinking water.\\
& (12) Q: What is Ross holding?\\ & \hspace{0.57cm}    A: Ross is holding a phone.\\

\hline
Level 2: Simple analysis
& (13) Q: What does Joey say to Chandler about the woman? \\ & \hspace{0.57cm} A: Joey says ``woman love babies and guys who love babies''. \\

& (14) Q: What did Joey and Chandler do? \\& \hspace{0.57cm}  A: drinking and dancing.\\
\hline
Level 3: Intermediate cognition
& (15) Q: How did Rachel regain the cart from the woman? \\ & \hspace{0.57cm}  A: Rachel get in a cart.\\ & (16) Q: How did Rachel figure out the truth of prom in high school? \\ & \hspace{0.57cm}  A: Rachel saw the video which shoot at the day.\\
\hline
Level 4: High-level reasoning
& (17) Q: Why does Rachel storm out of the office? \\ & \hspace{0.57cm}  A: She thinks the interviewer is trying to sleep with her.\\ & (18) Q: Why is Monica visiting Chandler at work?\\& \hspace{0.57cm}  A: They are going to look for houses together.\\
\hline
Level 5: Creative thinking
& (19) Q: If Ross did not break up with Rachel, what could have been the consequence? \\ & \hspace{0.57cm} A: They will love each other forever.\\
& (20) Q: If Rachel's parents didn't divorce, what could be different for her birthday party? \\& \hspace{0.57cm} A: Rachel's friends will organize the only one birthday party, not two separate parties. \\
\hline
\Xhline{3\arrayrulewidth}
\multicolumn{2}{l}{\textbf{Criterion 3: DIKW hierarchy}}\\
\noalign{\smallskip}
\hline
Level 1: Data-level
& (21) Q: What does Ross do?\\ & \hspace{0.57cm} A: Ross walks to the customer.\\
& (22) Q: Where does Phoebe grab the tissue?\\ & \hspace{0.57cm} A: Phoebe grabs from the kitchen.\\
\hline
Level 2: Information-level
& (23) Q: What is Rachel's emotion to Monica's situation?\\& \hspace{0.57cm} A: Rachel is horrified.\\
& (24) Q: What was Rachel doing before she comes into the room? \\& \hspace{0.57cm} A: She was dancing.\\
\hline

Level 3: Knowledge-level
& (25) Q: How do Joey and Chandler decide which baby to choose? \\& \hspace{0.57cm} A: Asking the transit authority employee.\\
& (26) Q: Why are Joey and Chandler at the Transit Authority \\& \hspace{0.57cm} A: To apply for a job.\\
\hline
Level 4: Wisdom-level
& (27) Q: Why now Rachel is telling Joey that her gynecologist tried to kill her?\\
& \hspace{0.57cm} A: To see if Joey could say a few words to her gynecologist too.\\
& (28) Q: If Chandler suffers from jet lag now,  what he could have done? \\& \hspace{0.57cm} A: He calls 911. \\
\Xhline{3\arrayrulewidth}
\end{tabular}
\end{center}
\end{table*}

\end{document}